\pdfoutput=1

\documentclass[11pt]{article}

\usepackage{EMNLP2023}

\usepackage{times}
\usepackage{latexsym}
\usepackage{algorithm}
\usepackage{algpseudocode}
\usepackage{amsmath}

\usepackage[T1]{fontenc}

\usepackage[utf8]{inputenc}

\usepackage{microtype}

\usepackage{graphicx}
\graphicspath{ {./images/} }

\usepackage{inconsolata}

\title{UGIF: UI Grounded Instruction Following}

\author{Sagar Gubbi Venkatesh \\
  Google \\
  \texttt{gubbi@google.com} \\\And
  Partha Talukdar \\
  Google \\
  \texttt{partha@google.com} \\\And
  Srini Narayanan \\
  Google \\
  \texttt{srinin@google.com} \\}

\begin{document}
\maketitle
\begin{abstract}
Smartphone users often find it difficult to navigate myriad menus to perform common tasks such as ``\emph{How to block calls from unknown numbers?}''. Currently, help documents with step-by-step instructions are manually written to aid the user. The user experience can be further enhanced by grounding the instructions in the help document to the UI and overlaying a tutorial on the phone UI. To build such tutorials, several natural language processing components including retrieval, parsing, and grounding are necessary, but there isn't any relevant dataset for such a task. Thus, we introduce UGIF-DataSet\footnote{pronounced with a soft-g: U-JIF}, a multi-lingual, multi-modal UI grounded dataset for step-by-step task completion on the smartphone containing 4,184 tasks across 8 languages. As an initial approach to this problem, we propose retrieving the relevant instruction steps based on the user's query and parsing the steps using Large Language Models (LLMs) to generate macros that can be executed on-device. The instruction steps are often available only in English, so the challenge includes cross-modal, cross-lingual retrieval of English how-to pages from user queries in many languages and mapping English instruction steps to UI in a potentially different language. We compare the performance of different LLMs including PaLM and GPT-3 and find that the end-to-end task completion rate is 48\% for English UI but the performance drops to 32\% for other languages. We analyze the common failure modes of existing models on this task and point out areas for improvement.
\end{abstract}

\section{Introduction}
\label{sec:intro}

Smartphone users often struggle to navigate the UI and get things done on the phone. This problem is particularly acute in developing countries due to varying literacy levels, high cost of phone ownership, etc. \citep{niuanthology}. Many of the frequently asked questions (FAQs) are documented on support sites\footnote{\url{https://support.google.com}} with step-by-step instructions describing what the user should do on the UI. We explore the problem of harnessing such help documents to create step-by-step tutorials overlaid on the UI of the phone.  

\begin{figure*}[!t]
    \centering
    \includegraphics[width=\linewidth]{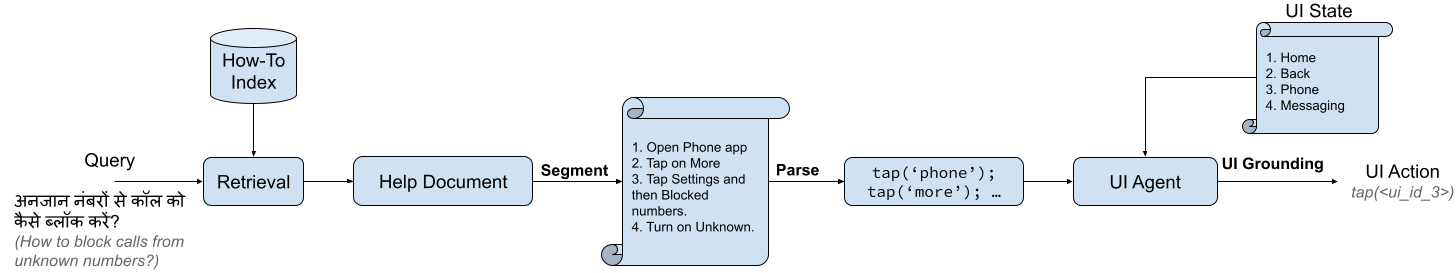}
    \caption{Our proposed initial approach retrieves the relevant help document, parses it to generates macros such as \texttt{tap()}, \texttt{toggle()}, \texttt{home()}, etc., and grounds each macro in the UI (Section~\ref{sec:intro}).}
    \label{fig:hero}
\end{figure*}



To create step-by-step tutorials on the UI using help documents, several natural language processing components including retrieval, parsing, and grounding are required. But no relevant dataset exists for this task in the multilingual setting. We build on prior work in the NLP community in this area \citep{li2020mapping} and extend it along the multilingual and multimodal directions. We collect a new multi-lingual, multi-modal UI grounded dataset called \texttt{UGIF-DataSet} to evaluate how well we can create step-by-step tutorials on the Android UI. It consists of 523 how-to queries per language and for each query, step-by-step instructions in English and a sequence of UI screenshots and actions that show how to complete the task. Each how-to query and UI sequence is available in 8 languages. An outline of the structure of this dataset is shown in Fig.~\ref{fig:dataset}. The data we release is focused on retrieval, parsing, and instruction following in Android, which should be of interest to the NLP community.

The central language related challenge here arises from the fact that many smartphone users are bilingual or even multilingual and frequently use non-EN languages. They ask queries in their native language, but the help documents are often available only in the English. Hence the need for cross-modal, cross-lingual retrieval. Furthermore, users may use a different UI / System language, and app developers do not always provide translations for every UI element resulting in some UI elements in English and the rest in the chosen system language. This necessitates cross-lingual UI grounding to map the instruction steps in English to UI screens containing different languages.

We propose an initial approach to this task that splits it into \textbf{retrieval}, \textbf{parsing}, and \textbf{grounding}. When the user utters a query, the matching FAQ page is retrieved from the support site using an off-the-shelf speech recognizer and a multi-lingual sentence embedding model \citep{feng2020language} to find the closest matching how-to question in the help document corpus. The step-by-step instructions in the help document are parsed using a large language model \citep{chowdhery2022palm} to generate macros such as \texttt{tap()}, \texttt{toggle()}, \texttt{home()}, etc. This macro sequence is used to create a tutorial on-device by grounding each macro in the UI using a multi-lingual sentence embedding model \citep{feng2020language} to find the closest matching UI element.

The contributions of this work are as follows:

\begin{itemize}
    \item We release \texttt{UGIF-DataSet}, a new multi-lingual, multi-modal dataset of how-to queries and sequences of UI screens and actions performed by human annotators. This is the first such multi-modal dataset of its kind.
    \item We evaluate parsing of step-by-step how-to instructions with large language models and UI grounding with multi-lingual BERT sentence embedding (LaBSE).
    \item Our results indicate that there is considerable room to improve performance, especially in non-English languages. Furthermore, we find that UI mismatches due to version changes as the app design evolves over time is a significant source of errors and presents both research and engineering challenges.
\end{itemize}

\section{Related Work}

\paragraph{Natural Language Instruction Following for UI navigation:} There have been several previous efforts at natural language conditioned UI navigation for desktop operating systems \citep{branavan2009reinforcement, branavan2010reading, xu2021grounding} and image editing applications such as Adobe Photoshop \citep{manuvinakurike2018edit}. More recently, there has been work on grounding natural language instructions to mobile user interfaces for automatically generating videos of help articles \citep{zhong2021helpviz}. Our work is an enhanced and updated successor to the \texttt{PixelHelp} dataset released in \citet{li2020mapping} with voice and text queries in eight languages, instruction steps in English, and UI screens in eight system languages.

\paragraph{Imitation learning and Reinforcement learning for UI navigation:} One can think of broadly two approaches to building a UI navigation agent: (a) scaling horizontally by building an agent that can handle a few simple tasks like searching for something, deleting an item, etc. that are useful across many different apps, and (b) scaling vertically by exposing a greater depth of functionality but only for a few applications. \citet{li2021learning} takes the former approach and uses behavior cloning and reinforcement learning to train agents for two specific skills: to install the specified app from the Play Store and another agent to find the search box in any app. To enable reinforcement learning research on Android UIs, \citet{toyama2021androidenv} introduces \texttt{AndroidEnv}, an open source platform for training RL agents. Similar to that, \texttt{WorldOfBits} is an open platform for training web navigation agents \citep{shi2017world, liu2018reinforcement}. In our work, we take the latter approach of exposing deeper functionality of a few popular apps by relying on help articles in the Android support site. We chose this because new users often ask goal oriented questions that require greater knowledge about how to navigate a particular app. Moreover, app developers often provide FAQs with common tasks in mind, so we can exploit the support pages to create UI grounded tutorials for new users.

\paragraph{Pre-training for UI tasks:} In the past few years, there has been a paradigm shift in deep learning towards pre-training on broad unlabelled datasets and fine-tuning on task specific data. \citet{bai2021uibert, he2021actionbert} pre-train a transformer model on a large number of screenshots obtained by crawling apps in smartphones in a manner similar to web crawling. Since our focus is on multilingual UI screens, we chose to use the pre-trained LaBSE \citep{feng2020language} for UI grounding, but utilizing broad UI data will be critical for future improvements. 

\paragraph{Large language models:} Large language models (LLMs) pre-trained on large corpora of text scraped from the web have shown remarkable few-shot generalization capability \cite{chowdhery2022palm, brown2020language}. We employ LLMs for parsing help articles but not for UI grounding since we prefer to do it on-device for privacy reasons.

\paragraph{Language grounding in human-robot interaction:} Language guided robot actions for human-robot interaction \citep{lynch2020language, venkatesh2021spatial} is a broadly related problem. However, taking actions on real robots is much more complex with uncertain outcomes, whereas precise actions can be performed on the UI with near certainty. As a result, the difficulty with UI grounded interactions is less about sensing and actuation and more about understanding user intent and navigating the app by understanding its structure using external resources such as support pages.


\paragraph{Icon and widget captioning:} Although Android allows developers to provide content description for images, not all app developers do so. To support a wide range of apps, it becomes necessary to recognize icons and widgets \cite{li2020widget, iconnet}. In our work, all the apps provide the necessary description, so icon captioning is not necessary.

\section{UGIF-DataSet: A New Multilingual Multimodal UI-grounded Instruction Following Dataset}
\label{sec:pixelhelp-plus-plus}

\begin{figure}[!t]
    \centering
    \includegraphics[width=\linewidth]{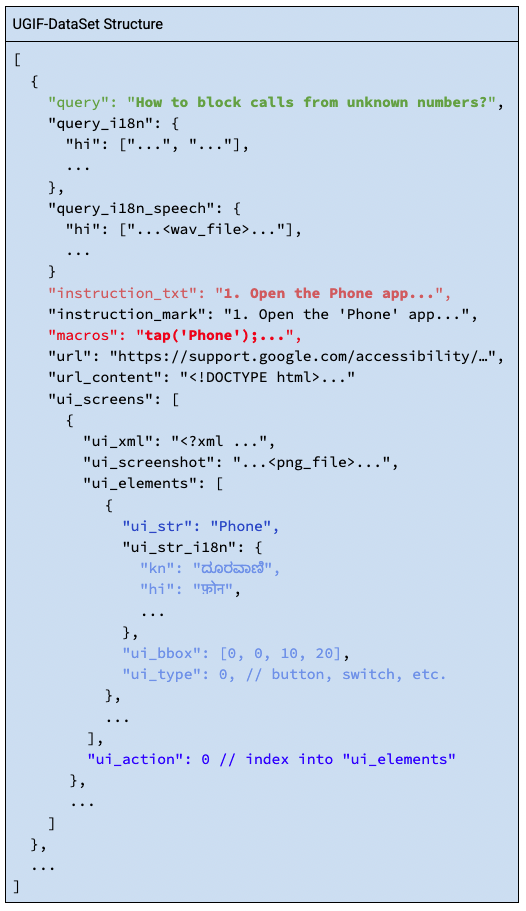}
    \caption{An outline of the \texttt{UGIF-DataSet} dataset, which consists of 523 pairs of how-to instructions and sequences of UI screens and actions (Section~\ref{sec:pixelhelp-plus-plus}).}
    \label{fig:dataset}
\end{figure}

Our goal is to build a UI navigation agent that can teach users how to perform tasks on the Android UI. To build such an agent and evaluate its performance, we collect a new multi-lingual, multi-modal UI grounded dataset called \texttt{UGIF-DataSet}\footnote{We will publicly release the dataset in the camera ready version.}.
It is a corpus of how-to queries in text and speech in multiple languages, instruction steps for each how-to paired with sequences of UI screens and actions as the how-to is completed by human annotators on Android devices with different UI language settings (Fig.~\ref{fig:dataset}).

The Pixel Help support pages provide step-by-step instructions for performing common tasks on Android. This is an example task: ``How to block unknown numbers?'' for which the instruction text is ``\emph{1. Open your Phone app 2. Tap More. 3. Tap Settings and then Blocked numbers. 4. Turn on Unknown}''. We crawl the Android support site and extract the how-to steps using simple rules that look for ordered lists under a header. Annotators\footnote{An internal tool was used for the annotation process following approval by the ethics, privacy, and legal committees.} translate and speak out loud the how-to query. They also parse the how-to steps to a sequence of macros in Table~\ref{table:macros}.

\begin{table}[!t]
\centering
\begin{tabular}{p{2cm}|p{4.8cm}}
\hline
\textbf{Macro} & \textbf{Function} \\
\hline
\texttt{tap($e$)} & Taps on the UI element specified in the argument ($e$) \\
\texttt{toggle($e$, val=True)} & Finds the UI element in the argument ($e$) and then searches for the nearest Switch element and taps on that \\
\texttt{home()} &  Presses the home button in Android \\
\texttt{back()} & Presses the back button \\
\texttt{prompt($a$)} & Requests the user to take some action ($a$) and waits until an action is performed \\
\hline
\end{tabular}
\caption{\label{table:macros}
List of all macros that can be generated from instruction steps (Section~\ref{sec:pixelhelp-plus-plus}).
}
\end{table}



For each how-to task, annotators are asked to operate a virtual Android device to carry out the steps in the how-to while the screen of the device and the annotator's actions are recorded. 
Just before each action taken by the annotator is forwarded to the virtual device and executed using  \texttt{UIAutomator} \cite{uiautomator}, we record a screenshot of the device, the view hierarchy in XML, and the action taken by the annotator at that step. We restrict the possible actions that the annotator can take at each step to: (a) tapping on a UI element, (b) pressing the home button, (c) pressing the back button, (c) prompting the end-user for an input, (d) toggling a switch / checkbox, (e) scrolling up / down, (f) noting the completion of the task, (g) noting an error in the how-to instruction text and ending the recording before completion.

The manual annotation process for collecting UI screens from the Android emulator scales linearly with the number of UI languages. To mitigate this, we collect UI screens from annotators only in English and search for each UI string in the resources directory of the app's APK and replace it with the translation provided by the developer in the APK wherever it is available. If a translation is unavailable, we default to English. A typical UI screen has a mixture of strings in English and other languages, but this is distinct from code mixing where two languages are used in a single sentence.  

The \texttt{UGIF-DataSet} dataset has 152 (train) / 106 (dev) / 265 (test) samples. It includes tasks in the following apps: Settings, Google One, Gmail, Play Store, Contacts, Messages, Chrome, Maps, Camera, Google Photos, Google Earth, and Files. \texttt{UGIF-DataSet} differs from the \texttt{PixelHelp} dataset \citep{li2020mapping} in the following ways. It:

\begin{itemize}
    \item Contains UI elements in seven non-English languages: Hindi, Kannada, Marathi, Gujarati, Bengali, Swahili, Spanish.
    \item Includes how-to instructions that need user input such as \emph{``Select the email you want to move to trash''}.
    \item It is a multi-modal dataset that includes not only the view hierarchy of the screens but also a screenshot at each step of the execution.
    \item Does not assume that the UI element is visible on the screen. The annotator is allowed to scroll and find the UI element referred in the instruction text. 
    \item Includes samples where the instruction text is outdated and does not correspond to the current version of the UI. In such cases, annotators can either adapt the instructions to the current UI or declare an error if they are unable to complete the task. 
\end{itemize}

\section{Model}

UGIF has three components: Retrieval, Parsing, and Grounding. Based on text or speech input, the most relevant how-to instruction in English is retrieved and then parsed to generate macros. These macros are executed on the Android device by grounding them in the UI (Alg.~\ref{alg:ugif}).

\begin{algorithm}
\caption{UGIF end-to-end description}\label{alg:ugif}
\begin{algorithmic}
\State \texttt{steps} $\gets \boldsymbol{retrieve\_howto}(\texttt{user\_query})$
\State \texttt{macros} $\gets \boldsymbol{parse}(\texttt{steps})$
\State \texttt{i} $\gets 0$
\While{\texttt{i} $< len(\texttt{macros})$}
    \State \texttt{macro} $\gets$ \texttt{macros[i]}
    \State \texttt{action} $\gets$ $\boldsymbol{ground}(\texttt{macro}, \texttt{screen})$
    \If{\texttt{action} $\ne$ SCROLL}
        \State \texttt{i} $\gets \texttt{i}+1$
    \EndIf
\EndWhile
\end{algorithmic}
\end{algorithm}

\paragraph{Retrieval}We use Google Cloud Speech\footnote{https://cloud.google.com/speech-to-text} as an off-the-shelf speech recognizer to convert speech to text. A multilingual sentence embedding model \citep{feng2020language} is used to obtain a vector corresponding to the query, which is then used to retrieve the most similar how-to by cosine similarity in the UGIF-DataSet corpus.

\label{sec:parsing}


\paragraph{Parsing}The parsing model takes how-to instructions and generates a sequence of macros (Table~\ref{table:example_macros}). We tried various language models such as PaLM \cite{chowdhery2022palm},  GPT-3 \cite{brown2020language}, T5 \cite{raffel2020exploring}, and UL2 \cite{tay2022unifying}) to generate the macro given the instruction text. 

\begin{table}[!t]
\centering
\begin{tabular}{p{3cm}|p{3.75cm}}
\hline
\textbf{Instruction text} & \textbf{Macro sequence} \\
\hline
Open the Phone app. Tap Recents. & \texttt{tap("Phone"); tap("Recents");} \\
\hline
Open the Settings app. Tap Network \& Internet. Turn off wi-fi. & \texttt{tap("Settings"); tap("Network \& Internet"); toggle("wi-fi", False);} \\
\hline
\end{tabular}
\caption{\label{table:example_macros}
Sample instructions and corresponding macro sequences (Section~\ref{sec:parsing}).
}
\end{table}

\label{sec:grounding}

\begin{figure*}[!t]
    \centering
    \includegraphics[width=0.65\linewidth]{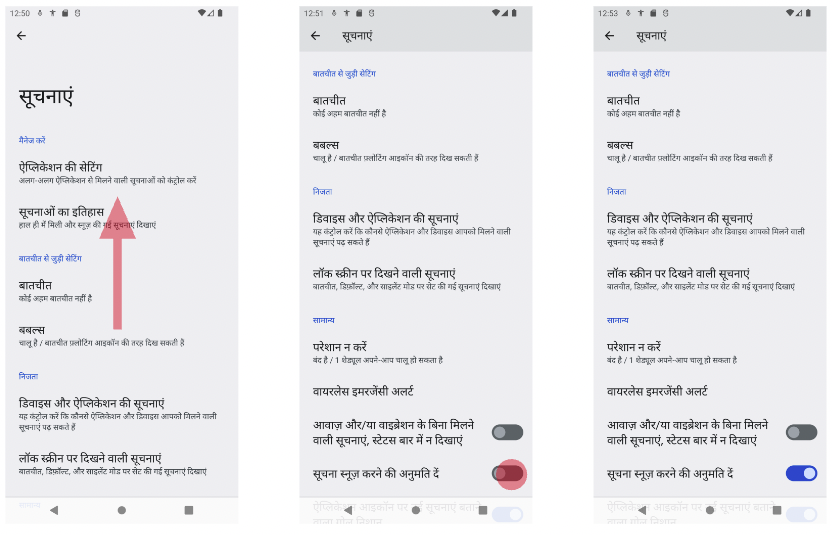}
    \caption{A sample sequence of UI screens and actions resulting from the execution of the macro: \texttt{toggle("Allow notification snoozing", True)}. The UI grounding model recognizes that none of the UI elements is a sufficiently close match to the string in the argument of the macro, scrolls down, finds a match, and taps on the nearest switch to turn it on (Section~\ref{sec:grounding}).}
    \label{fig:grounding_toggle}
\end{figure*}

\paragraph{Grounding}The grounding model takes a macro, potentially with arguments, as input along with the current UI screen and performs a series of actions on the UI to complete the task specified by the macro. The macros in our setup are described in Table~\ref{table:macros}.



For both \texttt{tap()} and \texttt{toggle()}, it is necessary to locate the UI element being referred to in the argument of these macros. i.e., we are given a macro with its argument referring to a UI element and a list of UI elements currently visible on the screen, and we must decide which element to pick (or to not pick at all and scroll for a better match). For finding the closest matching UI element, we experiment with jaccard similarity, UiBERT \citep{bai2021uibert}, and multi-lingual BERT sentence embedding (LaBSE) \citep{feng2020language}. The jaccard similarity between a UI element and the referring expression is measured by splitting the words in the UI string and the referring expression and finding the jaccard similarity between these two sets. The  LaBSE model generates embeddings for entire sentences, which we utilize to compute embeddings for each UI element and also for the input referring expression in the macro. The cosine similarity between the embeddings for the referring expression and the UI element is used as a scalar measure of the similarity between the arugment to the macro and the UI element. We use a scrolling threshold $T$ to decide whether to scroll or to accept a UI element currently on the screen. If the similarity metric is less than $T$, we choose to scroll down looking for a better match, whereas if the similarity metric is above $T$, the best matching UI element is chosen for interaction (either tapping or toggling). The appropriate value for $T$ is determined through experimentation on the development set. Likewise, we also use UiBERT to generate embeddings for all the UI elements on the screen along with the input referring expression, but with UiBERT we introduce an additional "Not found" UI element that the model is trained to choose if the scroll action is taken.

For the tapping macro, it is sufficient to look for the UI element most similar to the argument in the macro. However, for the toggle macro, when using LaBSE embeddings we first find the UI element referred to by the argument to the \texttt{toggle()} macro, and then look for an Android Switch element nearby in the view hierarchy (Fig.~\ref{fig:grounding_toggle}). This works as long as the app is using the standard Android Switch element and a straightforward XML layout of the mobile UI where the text field is close to the Switch element. Nevertheless, such heuristics are brittle and could be resolved by multimodal models which we leave for future work.

\section{Experiments}
\label{sec:experiments}

\begin{figure}[!t]
    \centering
    \includegraphics[width=\linewidth]{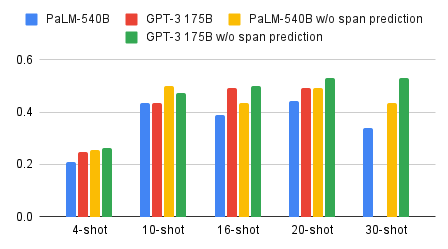}
    \caption{Parsing accuracy on the development set of \texttt{UGIF-DataSet} (Section~\ref{sec:parsing-results}).}
    \label{fig:parsing_dev_results}
\end{figure}

\begin{table}[t]
\centering
\begin{tabular}{lll}
\hline
\textbf{Model} & \textbf{Parsing}\\
\textbf{configuration} & \textbf{accuracy}\\
\hline
PaLM 540B 20-shot ICL & 46\% \\
GPT-3 175B 20-shot ICL & 50.9\% \\
PaLM 8B soft prompt tune & 49.1\% \\
PaLM 62B soft prompt tune & 64.9\% \\
PaLM 540B soft prompt tune & 66.8\% \\
UL2 20B full finetune & 66.8\% \\
T5 11B full finetune & 66.8\% \\
PaLM 8B full finetune & 64.5\% \\
PaLM 62B full finetune & 67.5\% \\
\textbf{PaLM 540B full finetune} & \textbf{70.1\%} \\
\hline
\end{tabular}
\caption{\label{table:parsing_results}
Parsing accuracy of pre-trained models on the \texttt{UGIF-DataSet} test set. In-context learning (ICL) is with 20 randomly selected training samples (single run). Fine-tuning and soft prompt-tuning with a 50-token soft prompt prefix (Section~\ref{sec:parsing-results}) is performed with all 158 training samples and hyper-paramter search over dropout values 0.0, 0.02, 0.05, 0.1, and 0.2 on 256 TPUs for about 24 hrs each. For fine-tuning, the best dropout was 0.1 with training for 10k steps, and for soft prompt-tuning, the best dropout was 0.0 with training for 17.5k steps.
}
\end{table}



\begin{table*}[!t]
\centering
\begin{tabular}{lllllllll}
\hline
\textbf{Model configuration} & \multicolumn{8}{c}{\textbf{UI Language}} \\
{}  & en & kn & mr & gu & hi & bn & es & sw \\
\hline
Oracle parse, Jaccard ground & 55.4 & --- & --- & --- & --- & --- & --- & --- \\
Oracle parse, UiBERT ground & 31.7 & --- & --- & --- & --- & --- & --- & --- \\
Oracle parse, LaBSE ground & 52.8 & 36.6 & 39.2 & 41.5 & 43.7 & 40.7 & 49.8 & 35.4 \\
PaLM 540B parse, LaBSE ground & 48.6 & 33.6 & 36.6 & 38.5 & 40 & 37.7 & 46.4 & 32.1 \\
\hline
\end{tabular}
\caption{\label{table:end_to_end_results}
End-to-end task completion success rate of different model configurations on the \texttt{UGIF-DataSet} test set (Section~\ref{sec:grounding-analysis}).
}
\end{table*}

The \texttt{UGIF-DataSet} dataset contains manually annotated oracle parses (macro sequences) for each how-to instruction text. We measure parsing accuracy by looking for an exact match between the generated parses and the oracle parses.

The dataset also contains manually annotated screen-action sequences for the entire how-to, but it does not have such sequences for each macro. So, to evaluate the grounding model, we consider the end-to-end task completion success rate. Although it is possible to complete each task in more than one way, we want to follow the how-to instruction text exactly, so we consider a task to be completed successfully only if the entire sequence of actions predicted by the model exactly matches the sequence of actions taken by the annotator.


\subsection{How well does retrieval work across languages?}
\label{sec:retrieval-results}

The multilingual sentence embedding model \citep{feng2020language} is excellent at matching how-to queries in non-EN languages to how-to queries in English (Table~\ref{sec:retrieval-results}). Examination of the failures with non-EN text queries revealed noise in the dataset where a small percentage of queries are repetitions with minor variations such as punctuation. When Google Cloud Speech API is used as an off-the-shelf automated speech recognizer (ASR) to convert speech input to text, there is a measurable drop in performance across all languages, but the reduction is large for Swahili. We also noticed that ASR failures were due to poor voice clarity, background noise, and more common with technical terms such as "cache".

\begin{table}[t]
\centering
\begin{tabular}{lll}
\hline
\textbf{Query} & \textbf{Oracle text} & \textbf{ASR text} \\
\textbf{Language} & \textbf{P@1} & \textbf{P@1} \\
\hline
en & 100 & 94.4 \\
kn & 97.9 & 88.6 \\
mr & 98.1 & 91.7 \\
gu & 97.3 & 89.6 \\
hi & 94.6 & 91.3 \\
bn & 97.3 & 91.2 \\
sw & 93.0 & 76.4 \\
es & 96.5 & 94.8 \\
\hline
\end{tabular}
\caption{\label{table:retrieval_results}
Comparison of performance for retrieving the closest matching how-to in English from queries in different languages (Section~\ref{sec:retrieval-results}).
}
\end{table}

\subsection{How does parsing performance scale with dataset and model size?}
\label{sec:parsing-results}



There is a steep increase in parsing performance from 4-shot prompting to 10-shot prompting (Fig.~\ref{fig:parsing_dev_results}). At 30 examples, the number of tokens in the input exceeds the maximum that the model can handle and performance deteriorates. Marking salient spans in the instruction text as an intermediate step for chain of thought prompting \citep{wei2022chain} degrades parsing performance. When all the available training samples are used with full fine-tuning or soft prompt tuning \citep{lester2021power}, the resulting performance is significantly better than few-shot prompting (Table~\ref{table:parsing_results}). The parsing accuracy increases only modestly with model size when full fine-tuning is used. However, with soft prompt tuning, there is more benefit to using larger models.

\subsection{What are the common failure modes of large language models for parsing?}
\label{sec:parsing-analysis}

\begin{figure}[!t]
    \centering
    \includegraphics[width=\linewidth]{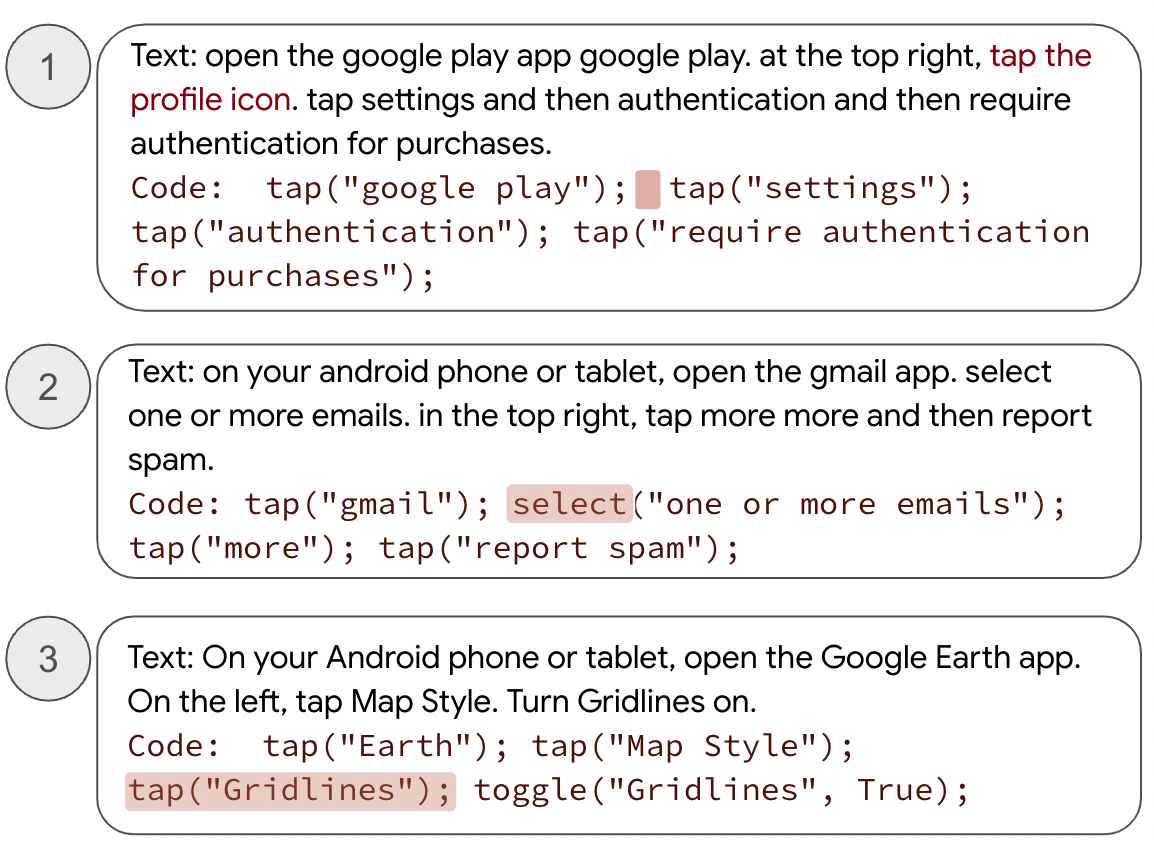}
    \caption{Incorrect sequences of macros generated by the 20-shot prompted  PaLM 540B model. In the first example, the macro \texttt{tap("profile icon")} is omitted in the output. In the second example, the model hallucinates the non-existent \texttt{select()} macro. In the last example, it has generated an un-necessary tap: \texttt{tap("Gridlines")} (Section~\ref{sec:parsing-analysis}). }
    \label{fig:parse_errors}
\end{figure}

\begin{figure}[!t]
    \centering
    \includegraphics[width=0.8\linewidth]{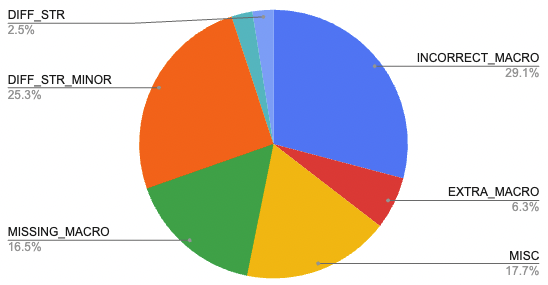}
    \caption{The types of parsing errors made by the PaLM 540B finetuned model (Section~\ref{sec:parsing-analysis}).}
    \label{fig:parse_errors_pie_chart}
\end{figure}

We examined the test samples where the model's predictions were incorrect (Fig.~\ref{fig:parse_errors}) and found the PaLM 540B finetuned model (a) generated incorrect macros, (b) made minor errors in predicting the span of the argument such as including the full stop, (c) missed salient parts of the input instruction resulting in skipped macros, and (d) hallucinated non-existent macros (Fig~\ref{fig:parse_errors_pie_chart}).

\subsubsection{How well do existing models work for UI grounding?}
\label{sec:grounding-analysis}



\begin{figure}[!t]
    \centering
    \includegraphics[width=\linewidth]{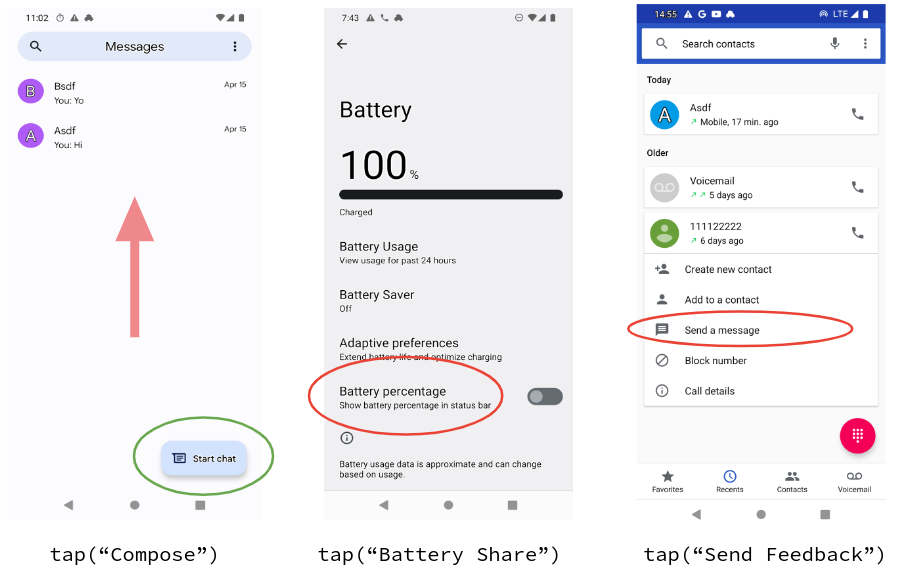}
    \caption{The UI grounding model chooses incorrect actions given the UI state and the macro. In the first example, the model should have tapped on ``Start chat'' as the matching element for ``Compose'' but instead tries scrolling down and throws an error that a matching UI element is not found. In the second example, the model should have scrolled down to find ``Battery share'' but instead erroneously selects the partially matching ``Battery percentage''. In the last example, the model should have recognized that the ``Send feedback'' button is missing in the UI and thrown an error, but instead erroneously selects the partially matching ``Send a message'' button (Section~\ref{sec:experiments}). }
    \label{fig:ground_errors}
\end{figure}

\begin{figure}[!t]
    \centering
    \includegraphics[width=0.78\linewidth]{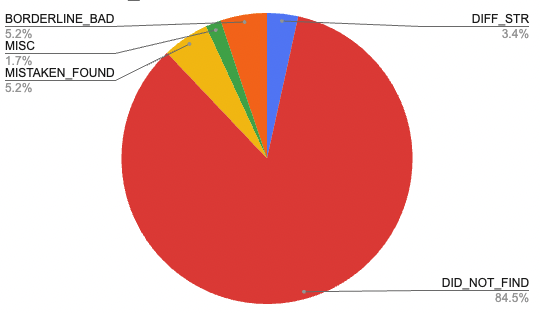}
    \caption{Categories of UI grounding errors using LaBSE (Section~\ref{sec:grounding-analysis}).}
    \label{fig:ground_errors_pie_chart}
\end{figure}

We find that even simple string matching models can offer good performance when the language in the how-to matches the UI language (Table.~\ref{table:end_to_end_results}). To our surprise, UiBERT underperformed this baseline. When the instruction text and the UI language are different, we have to use LaBSE which is a multilingual model, but we find that performance with English is still better than other languages. An examination of the incorrectly predicted samples (Fig.~\ref{fig:ground_errors}) using LaBSE revealed these modes of failure (Fig.~\ref{fig:ground_errors_pie_chart}): (a) Inexact string matching fails and the model keeps scrolling in the hope of a better match which it never finds (84.5\%), (b) the model overtriggers and chooses an inexact match instead of scrolling and looking for a better match (5.2\%), (c) the model lacks knowledge of common UI patterns and app names, so it gets confused between ``Play Store'' and ``Google One'' when trying find the closest match for ``Google Play'' (5.2\%).

The cases where the grounding model overtriggers and chooses a partially matching UI element and fails to either scroll down or recognize that the how-to is outdated results in incorrectly executed steps on the UI. These are of the most serious concern since they lead to a poor user experience. Moreover, help articles frequently become out-of-date as evidenced by the fact that ~29\% of the samples in \texttt{UGIF-DataSet} are marked by annotators as having instruction text not matching the UI in Android 12.

We also evaluated our best performing model on the \texttt{PixelHelp} dataset \citep{li2020mapping}. Table~\ref{table:e2e_comparison} shows that \texttt{UGIF-DataSet} is a harder dataset with significantly greater headroom for improvement especially in non-EN languages.




\begin{table}[t]
\centering
\begin{tabular}{ll}
\hline
\textbf{Model, Dataset} & \textbf{Success}\\
 & \textbf{rate} \\
\hline
\citet{li2020mapping}, PixelHelp (en) & 70.5\% \\
Ours, PixelHelp (en) & 71.1\% \\
Ours, UGIF-DataSet (en) & 48.6\% \\
Ours, UGIF-DataSet (sw) & 32.1\% \\
\hline
\end{tabular}
\caption{\label{table:e2e_comparison}
Comparison of our best performing model (PaLM 540B for parsing and LaBSE for grounding) on different datasets. There is a wide gap between the model performance on the \texttt{PixelHelp (en)} dataset and \texttt{UGIF-DataSet (sw)} which suggests considerable headroom for improvement (Section~\ref{sec:experiments}).
}
\end{table}

\section{Conclusion}



We proposed helping new smartphone users by showing them how to perform tasks on the UI based on voice queries. We evaluated existing language and sentence similarity models for the task of retrieving and executing how-to instructions on the UI where the UI language potentially differs from the language used in the instruction text. The models we build for this task must be capable of adapting to minor variations in the UI as the newer versions of the app are frequently released and instructions become outdated.  Multilingual UIs pose the challenge of having to simultaneously work with multiple languages in a single UI screen since app developers may not have provided translations for all UI elements. Finally, our evaluation of current pre-trained models suggests that there is significant room for improvement and that a multimodal language-UI foundation model could lead to substantial gains.

\section{Limitations}

The UGIF-DataSet contains only one speech sample per query in each language, so the diversity of speech samples is limited. All the instructions have been scraped from the Google support site, so our evaluation of parsing does not cover instruction text on forums and other support sites. All the UI captures in our dataset start at the home screen, but it would be desirable to also evaluate UI grounding from arbitrary starting points.



\section{Ethical Considerations}

Automated agents that operate over the UI could potentially be misused and pollute the global digital commons by making it harder for app developers to trust that the user is a real user. As a result, it is possible that many developers may choose to mitigate this by requiring some form of identification to use the app, which could hurt marginalized communities and users who struggle with such entry barriers. Further investigations and user studies on the benefits of automated UI agents will be helpful.

\bibliography{anthology,custom}
\bibliographystyle{acl_natbib}

\end{document}